\pdfoutput=1

\documentclass[11pt]{article}

\usepackage[final]{acl}

\usepackage{times}
\usepackage{placeins}
\usepackage{latexsym}
\usepackage{float}
\usepackage[T1]{fontenc}

\usepackage[utf8]{inputenc}

\usepackage{microtype}

\usepackage{inconsolata}

\usepackage{graphicx}
\usepackage{amsmath}
\usepackage{algorithm}
\usepackage{algpseudocode}
\usepackage{amsmath}
\usepackage{array}
\usepackage{booktabs}
\usepackage{mathtools}

\usepackage{titletoc}
\usepackage[toc,page,header]{appendix}
\usepackage{minitoc}

\usepackage{amsfonts}

\newcommand{\Revision}[1]{\textcolor{black}{#1}}
\title{Effective Skill Unlearning through Intervention and Abstention}

\author{Yongce Li \\
  UCSD HDSI \\
  \texttt{yol013@ucsd.edu} \\\And
  Chung-En Sun \\
  UCSD CSE\\
  \texttt{cesun@ucsd.edu} \\\And
  Tsui-Wei Weng \\
  UCSD HDSI\\
  \texttt{lweng@ucsd.edu} \\}

\begin{document}
\maketitle
\begin{abstract}
Large language Models (LLMs) have demonstrated remarkable skills across various domains. Understanding the mechanisms behind their abilities and implementing controls over them is becoming increasingly important for developing better models. In this paper, we focus on skill unlearning in LLMs, specifically unlearning a particular skill while retaining their overall capabilities. We introduce two lightweight, training-free machine skill unlearning techniques for LLMs. First, we observe that the pre-activation distribution of neurons in each Feed-Forward Layer (FFL) differs when the model demonstrates different skills. Additionally, we find that queries triggering the same skill cluster within the FFL key space and can be separated from other queries using a hypercube. Based on these observations, we propose two lightweight, training-free skill unlearning methods via \textit{intervention} and \textit{abstention} respectively: \texttt{Neuron Adjust} and \texttt{Key Space Detection}. We evaluate our methods on unlearning math-solving, Python-coding, and comprehension skills across seven different languages. The results demonstrate their strong unlearning capabilities for the designated skills. Specifically, \texttt{Key Space Detection} achieves over 80\% relative performance drop on the forgetting skill and less than 10\% relative performance drop on other skills and the model's general knowledge (MMLU) for most unlearning tasks. \footnote{Our code is available at \href{https://github.com/Trustworthy-ML-Lab/effective_skill_unlearning}{https://github.com/Trustworthy-ML-Lab/effective\_skill\_unlearning}}

\end{abstract}

\doparttoc 
\faketableofcontents 

\begin{table*}[h]
\small
\centering
\scalebox{1.0}{
\begin{tabular}{@{}l||cc|cc|c@{}}
\toprule
 & \multicolumn{2}{c}{(I) Efficiency} & \multicolumn{2}{c}{(II) Performance} & \multicolumn{1}{c}{(III) Scalability} \\ \midrule
Method: & \begin{tabular}[c]{@{}c@{}} Does not \\ require training \end{tabular} & \begin{tabular}[c]{@{}c@{}} No Inference \\ time cost \end{tabular} & \begin{tabular}[c]{@{}c@{}} High quality \\ unlearning\end{tabular} & \begin{tabular}[c]{@{}c@{}}Maintain model \\ overall capability\end{tabular} & \begin{tabular}[c]{@{}c@{}}Applicable to \\ large models \end{tabular}\\ \midrule
Retrain from scratch & No & \textbf{Yes} & \textbf{Yes}  & \textbf{Yes} & No  \\
Fine-tuning based unlearning & No & \textbf{Yes} & \textbf{Yes} & No & \textbf{Yes}   \\
In-context unlearning & \textbf{Yes} & \textbf{Yes} & No & \textbf{Yes} & \textbf{Yes}  \\
\texttt{Selective Pruning} & \textbf{Yes} & \textbf{Yes} & \textbf{Yes} & No & \textbf{Yes}  \\ \midrule
\texttt{Neuron Adjust} (\textbf{Ours}) & \textbf{Yes} & $O(1)$ & \textbf{Yes} & \Revision{Partial} & \textbf{Yes} \\
\texttt{Key Space Detection} (\textbf{Ours}) & \textbf{Yes} & $O(1)$ & \textbf{Yes} & \textbf{Yes} & \textbf{Yes} \\ \bottomrule
\end{tabular}
}
\vspace{-10pt}
\caption{Comparison of our method against existing machine unlearning methods, including retraining, fine-tuning based methods, in-context unlearning, and a prune-based unlearning method \texttt{Selective Pruning} \cite{pochinkov2023dissecting}.
}
\label{tab:pros_cons}
\end{table*}

\section{Introduction}


In recent years, the superior capabilities demonstrated by Large Language Models (LLMs) have attracted significant research interest. Without training on task-specific datasets, LLMs exhibit strong skills in various domains such as math \cite{wei2022chain, imani-etal-2023-mathprompter, cobbe2021training}, coding \cite{austin2021program, li2022competition}, and language comprehension \cite{shi2023language}. Understanding the mechanisms behind these abilities and implementing controls over them are becoming increasingly important for developing stronger, safer, and more interpretable models.

\begin{figure}[t] 
  \includegraphics[width=\columnwidth] {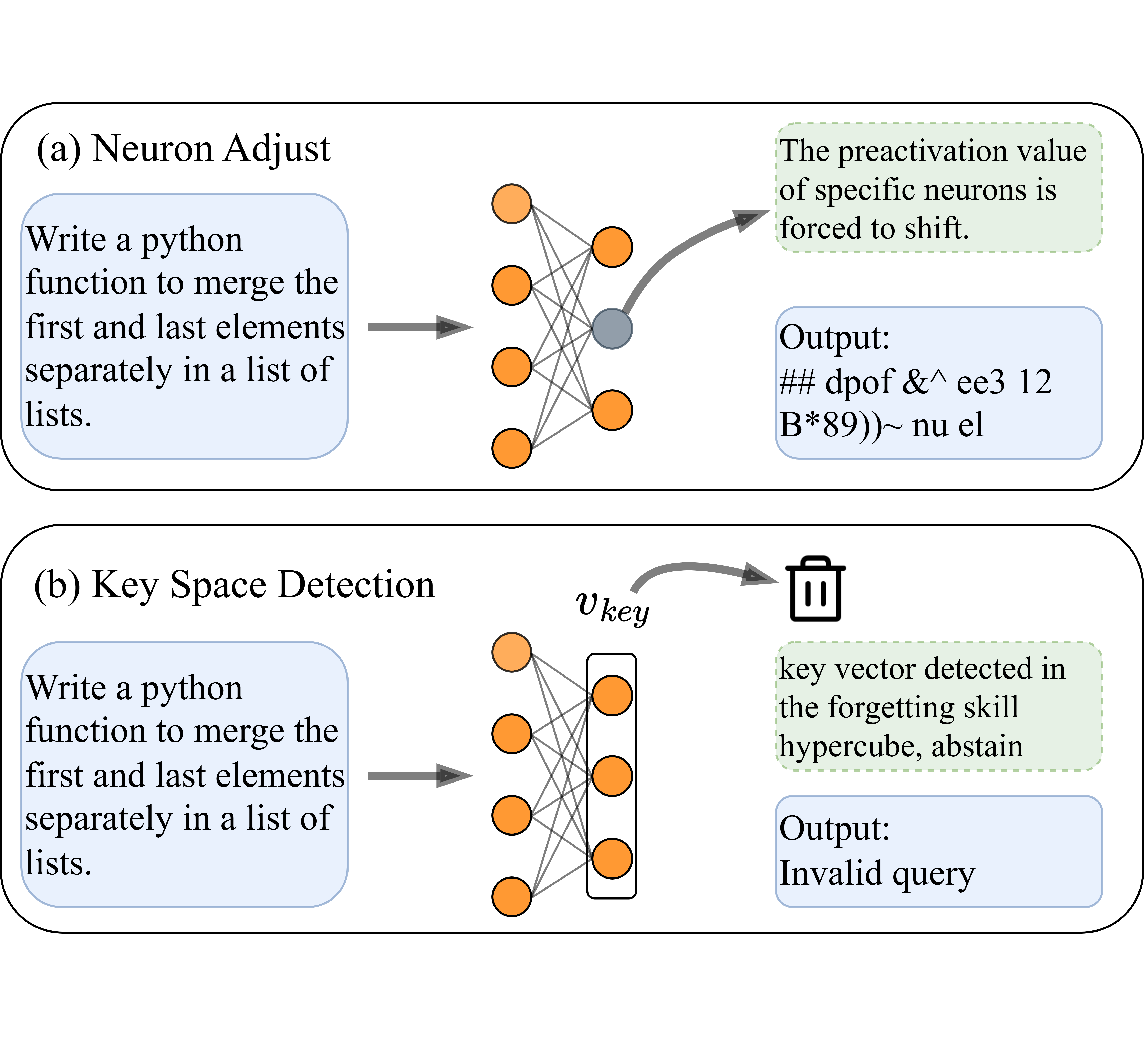}
  \caption{An overview of the proposed skill unlearning methods: \texttt{Neuron Adjust} (through \textit{intervention}) and \texttt{Key Space Detection} (through \textit{abstention}). This example illustrates forgetting coding skill.}
  \label{fig:overview}
\end{figure}

A recent line of research focuses on machine unlearning \cite{yao2023large, liu2024rethinking}, which aims to remove the knowledge LLMs have acquired from specific datasets while maintaining their causally unrelated knowledge. In this work, we focus on a variant of machine unlearning called skill unlearning, which aims to remove a specific skill (e.g., coding skill, elementary math-solving skill) from the LLM while retaining its other skills. Skill unlearning helps researchers control certain behaviors of LLMs, providing insights into when and how a model demonstrates a particular skill.

Currently, most unlearning methods \cite{lu2022quark, jang-etal-2023-knowledge, wang-etal-2023-kga, yu-etal-2023-unlearning, eldan2023whos, chen-yang-2023-unlearn, yao2023large} rely on fine-tuning, which becomes increasingly costly as LLMs grow larger. Other unlearning methods \cite{wu-etal-2023-depn, pochinkov2023dissecting} involve pruning dataset-related sets of neurons, which we show can harm the model's overall capabilities. In this paper, we introduce two new machine skill unlearning methods that are \textit{training-free} and have minimally impact the model's overall capabilities. We first observe that feed-forward layer neurons exhibit different pre-activation distributions when the model demonstrates different skills. Based on this observation, in section \ref{sec:neuron_adjust} we propose \texttt{Neuron Adjust}, which probabilistically shifts neuron pre-activation values to retain the desired skill distribution during inference through \textit{intervention}. By considering the correlation among neurons, we further observe that neuron activation vectors cluster within different hypercubes in the feed-forward layer’s key space when the model demonstrates certain skills. Building on this, we introduce \texttt{Key Space Detection (KSD)} in section \ref{sec:description_of_KSP_idea}, which detects and blocks specific skill-related activations in the key space through \textit{abstention} by preventing query vectors from accessing the skill-specific hypercube.

Our contributions can be summarized as follows:
\begin{enumerate}
\item Motivated by the shift in neuron pre-activation distributions and the modularity of skill-triggering queries in the feed-forward layer's key space, we propose two novel machine skill unlearning methods, \texttt{Neuron Adjust} and \texttt{Key Space Detection}, which are scalable, training-free, and maintain the model's overall capabilities with minimal degradation.

\item Our experiments on math, code, and language skill unlearning demonstrate the effectiveness of the proposed two methods with $>80\%$ relative performance drop on the target forgetting skill and $<10\%$ drop on the model's general knowledge (MMLU \cite{hendryckstest2021}) and other skills. Specifically, \texttt{Key Space Detection} achieves nearly perfect skill unlearning with negligible overall capability drop.
\end{enumerate}
\Revision{Table \ref{tab:pros_cons} compares our methods with traditional unlearning methods and the skill unlearning method \texttt{Selective Pruning} \cite{pochinkov2023dissecting} across the dimensions of Efficiency, Performance, and Scalability.}
%

\begin{figure*}[t] 
  \includegraphics[width=1\linewidth]{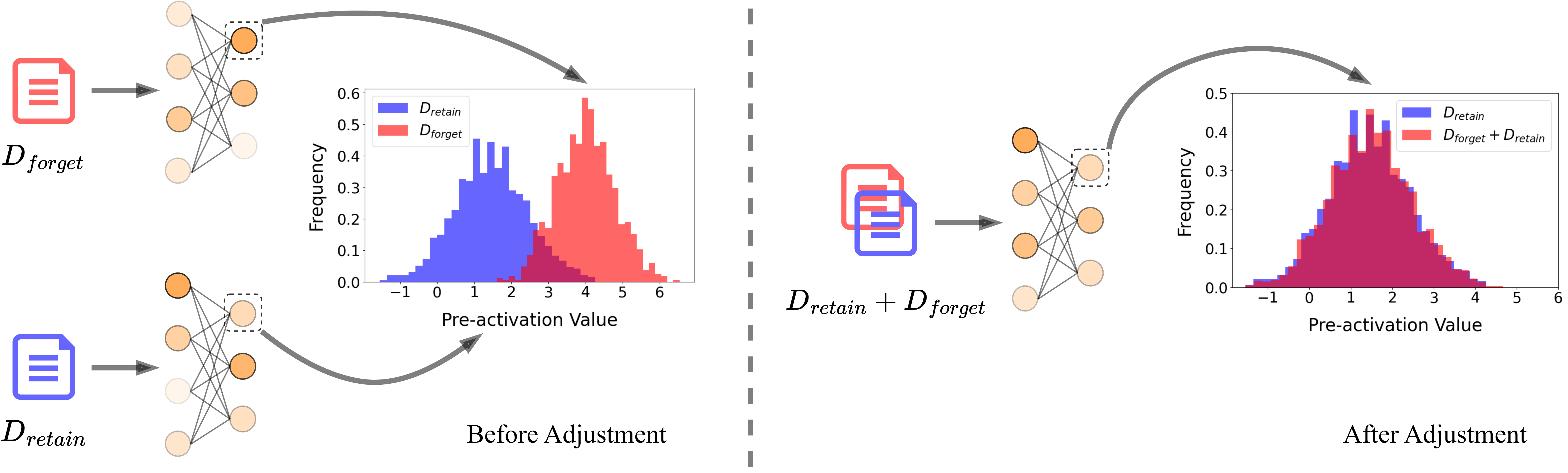}
  \vspace{-20pt}
  \caption{An overview of the \texttt{Neuron Adjust} method in section \ref{sec:neuron_adjust}. Before adjusting a neuron, the neuron has different distributions under the forgetting and retaining datasets. During inference time, \texttt{Neuron Adjust} algorithm will edit neurons with large distribution shift such that its pre-activation distribution will be close to the retaining distribution.}
  \label{fig:neuron_adjust}
\end{figure*}

\section{Related Work \& Background}

\textbf{Large language model machine unlearning.} This line of work aims to remove the influence of specific data points and the corresponding model capabilities without retraining the model from scratch. Most previous works on machine unlearning have focused on fine-tuning-based approaches \cite{lu2022quark, jang-etal-2023-knowledge, wang-etal-2023-kga, yu-etal-2023-unlearning, eldan2023whos, chen-yang-2023-unlearn, yao2023large}, which become increasingly costly as models grow larger. \citet{pawelczyk2024incontext} introduced in-context unlearning, which provides contextual inputs to the language model during the inference stage. Despite its cost-efficiency, it lacks unlearning quality and is difficult to generalize to large-scale unlearning.

Other training-free approaches focus on pruning or removing specific sets of behavior-related neurons in the model. \texttt{DEPN} \cite{wu-etal-2023-depn} is a pruning-based unlearning approach that removes neurons based on their cumulative privacy gradient. \texttt{Selective Pruning} \cite{pochinkov2023dissecting} is another pruning-based method, which removes neurons based on their relative importance to the forgetting dataset and the retaining dataset. However, the extent to which pruning-based methods affect the model's overall capabilities remains unknown and unjustifiable.

\noindent\textbf{Feed-forward neuron interpretability.} This line of work focuses on the interpretation of individual neurons, meaning that individual neurons represent meaningful concepts, both in vision models \cite{bau2020units, hernandez2022natural, oikarinen2023clip} and language models \cite{bills2023language, lee2023importance, cbllm}. Recent works have shown that neurons exhibit multisemanticity \cite{elhage2022superposition, bricken2023monosemanticity, huben2024sparse}, with some being expressible as a linear combination of concepts \cite{oikarinen2024linear}. By considering neuron activation vectors, we can also treat LLM's feed-forward layers (FFLs) as key-value memories, with neuron activation vectors as keys and the output of the FFLs as values \cite{geva-etal-2021-transformer, meng2022locating}.

\noindent\textbf{Unlearning settings.}
In this paper, we focus on unlearning a specific skill or capability of a language model while retaining another. In the following sections, we denote $D_{retain}$ as the dataset capturing the skill we want the model to retain performance, and $D_{forget}$ as the dataset capturing the skill we want the model to forget.

\Revision{Our methods involve} operations on the FFL in large pretrained autoregressive transformer decoder models, which takes the layer-normed input $z \in \mathbb{R}^H$ from the residual stream:
\begin{equation}
\begin{split}
    \texttt{FFL}^{(l)}(z) = W_{\text{down}}^{(l)}\sigma \left(W_{\text{up}}^{(l)}z\right) \nonumber,
\end{split}
\end{equation}
where $z$ is first mapped to a higher-dimensional space by an up-projection linear transformation $W_{\text{up}}^{(l)}$ and a non-linear activation function $\sigma$ to obtain neuron activations, and then mapped back to $\mathbb{R}^H$ space with a down-projection linear transformation $W_{\text{down}}^{(l)}$. Modern LLMs also utilize gated linear units (GLUs) in FFLs. Instead of directly activating each neuron, GLUs use a gating mechanism to control the information flow of each neuron:
\begin{equation}
\begin{split}
    \texttt{FFL}^{(l)}(z) = W_{\text{down}}^{(l)}\left(\sigma \left(W_{\text{gate}}^{(l)}z\right) \odot W_{\text{up}}^{(l)}z\right) \nonumber,
\end{split}
\end{equation}

where $\odot$ is element-wise vector multiplication.
In the following sections, we consider $W_{\text{up}}^{(l)}z$ in traditional FFL and $W_{\text{gate}}^{(l)}z$ in GLU-FFL as the neuron pre-activations, and vectors after activation function as the key vectors, i.e. we have $v^{(l)}_{key} = \sigma \left(W_{\text{up}}^{(l)}z\right)$ and $v^{(l)}_{key} = \sigma\left(W_{\text{gate}}^{(l)}z\right) \odot W_{\text{up}}^{(l)}z$ in regular FFL and GLU-FFL respectively.


\begin{figure*}[h]
  \includegraphics[width=1\linewidth]{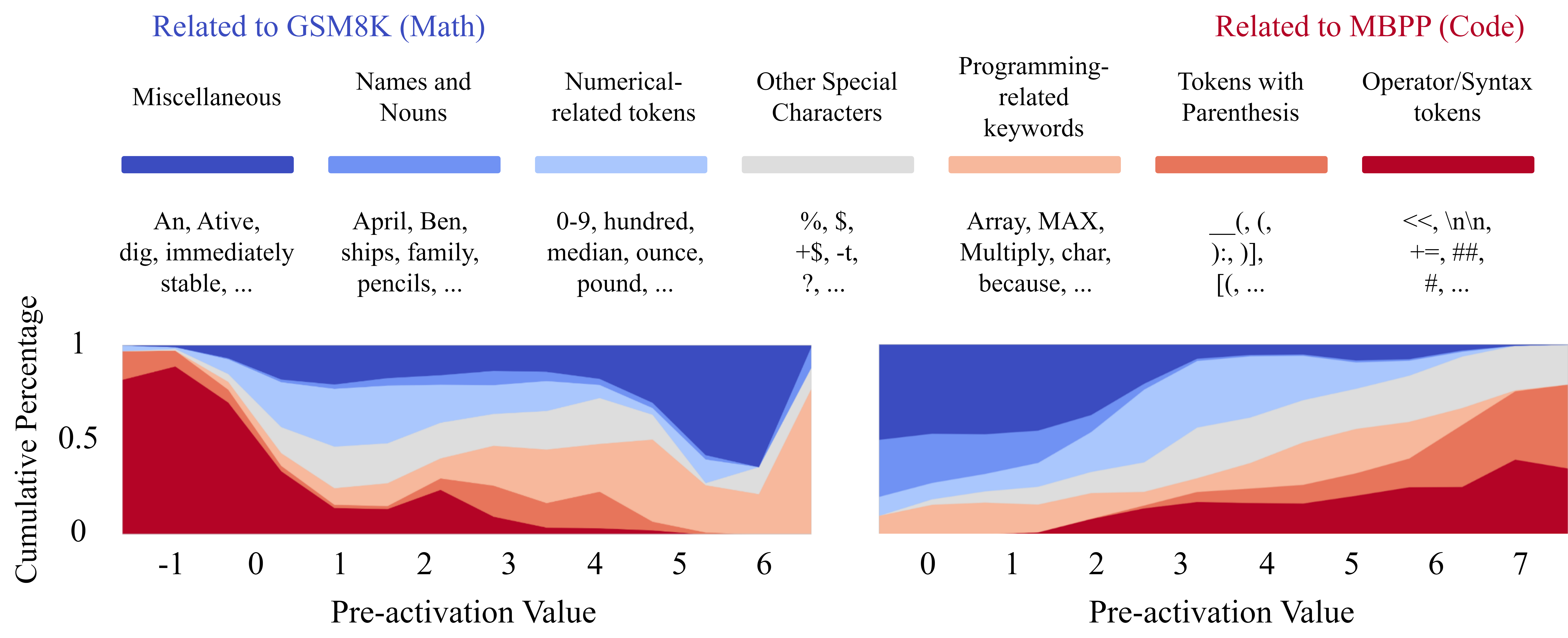}
  \vspace{-20pt}
  \caption {Category distribution over different pre-activation value ranges for neuron at layer 17, index 693 (left), and layer 0, index 13366 (right).}
  \label{fig:pre_activation_distribution}
\end{figure*}

\begin{figure*}[h] 
  \includegraphics[width=1\linewidth]{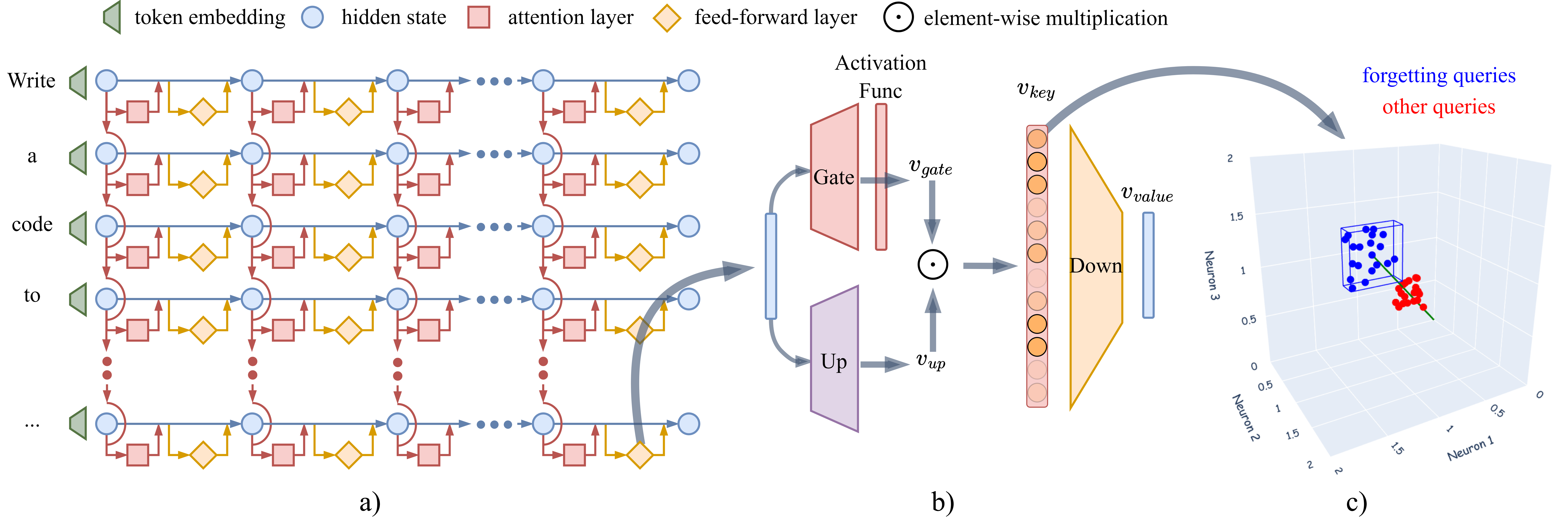}
  \vspace{-20pt}
  \caption {Overview of the \texttt{Key Space Detection} method in section \ref{sec:description_of_KSP_idea}. a) shows the structure of decoder-based large language models. b) shows the components of a GLU-based feed-forward layer in the LLM, where $v_{\text{key}}$ is located in the key space we aim to prune. c) is an example of a 3-neuron key space. The blue hypercube is formed by $\{\vec\mu_{D} \pm \alpha\vec\sigma_{D}\}$, where $\vec\mu_{D}$ and $\vec\sigma_{D}$ are the sample mean vector and standard deviation vector of $v_{\text{key}}$ when probing the model with the forgetting dataset. During every inference step, if we detect $v_{\text{key}} \in \{\vec\mu_{D} \pm \alpha\vec\sigma_{D}\}$, we prohibit the model from generating the output. 
  }
  \label{fig:MU_hyperrectangle}
\end{figure*}

\section{Inference Time Neuron Adjustment}
\label{sec:neuron_adjust}
In this section, we introduce \texttt{Neuron Adjust}, a post-hoc, training-free machine unlearning technique for large language models achieved by inference time neuron pre-activation value adjustment. An overview of \texttt{Neuron Adjust} method is shown in Figure \ref{fig:neuron_adjust}. In section \ref{sec:motivation_neuron_adjust}, we show the motivation of the method that some neurons' pre-activation distributions differ when the model demonstrates different capabilities. Based on the observation, we describe the \texttt{Neuron Adjust} method in section \ref{Neuron Adjust Algorithm}.

\subsection{Case Study: Neuron Pre-Activation Distribution Shift}
\label{sec:motivation_neuron_adjust}


We perform a case study to show how neuron pre-activation distribution changes when the model demonstrates math and coding skills separately.  We choose GSM8K \cite{cobbe2021training} and MBPP \cite{austin2021program} as the two datasets that characterize the math and Python coding skills of the model, and \texttt{Gemma-2b-it} \cite{gemmateam2024gemma} as the subject model to study. We probe the model with the two datasets, and document each neuron's (token, pre-activation) pair, and then prompt GPT-4 \cite{openai2024gpt4} to categorize tokens into meaningful categories as in Figure \ref{fig:pre_activation_distribution}. For the two neurons, although they are highly activated by "Programming-related keywords," "Operator/Syntax tokens," and "Tokens with Parentheses," they both show positive activations for "Numerical-related tokens." This case study demonstrates that neurons are multi-functional, and simply pruning them would be harmful to the model's overall capabilities.

\subsection{\texttt{Neuron Adjust} Algorithm}
\label{Neuron Adjust Algorithm}

Based on the observation that neuron pre-activations exhibit different distributions when the model demonstrates different skills and the polysemantic property of neurons, we propose \texttt{Neuron Adjust}, a probabilistic skill unlearning technique applied to the subject model during inference time. \texttt{Neuron Adjust} unlearns one skill of the model while retaining another skill by shifting each neuron's pre-activation from the forgetting skill distribution to the retaining skill distribution. Algorithm \ref{algo:neuron_adjust} shows the pseudo-code of \texttt{Neuron Adjust}, which mainly consists of two parts:

\begin{algorithm}[h!]
\caption{\texttt{Neuron Adjust} Algorithm for neuron $n_i$ and inference time pre-activation $v$}
\begin{algorithmic}[1]
\State \textbf{Input:} $D_{retain}$, $D_{forget}$, $v$

\State Probing the model with $D_{retain}$ and $D_{forget}$, approximate sample mean and std:
 $$n_i | D_{retain} \sim \mathcal{N}(\mu_r, \sigma_r)$$
 $$n_i | D_{forget} \sim \mathcal{N}(\mu_f, \sigma_f)$$

\State Calculate $p_r = P(v | \mathcal{N}(\mu_r, \sigma_r))$
\State Calculate $p_f = P(v | \mathcal{N}(\mu_f, \sigma_f))$

\If{$p_r < p_f$}
    \State {$\alpha \gets \frac{p_f}{p_r+p_f}$}
    \State {$v_{\text{adjust}} \gets 2\mu_r - \left(\frac{v - \mu_f}{\sigma_f} \sigma_r + \mu_r \right)$ with \indent probability $\alpha$}
    \State {$v_{\text{adjust}} \gets v$ with probability $1-\alpha$}
\Else
    \State {$v_{\text{adjust}} \gets v$}
\EndIf
\State \textbf{Output:} $v_{\text{adjust}}$
\end{algorithmic}\label{algo:neuron_adjust}
\end{algorithm}

\begin{enumerate}
    \item Probe the model with the forgetting and retaining datasets. For each neuron, assume that the forgetting and retaining pre-activation distributions follow a normal distribution. Approximate the means and standard deviations (stds) of these two distributions using sample pre-activation values.
    \item During inference, when a neuron has a pre-activation value $v$, calculate the likelihood of $v$ being sampled from each of the two distributions. If $v$ is more likely to be sampled from the retaining distribution, keep the value. Otherwise, shift $v$ towards the retaining distribution. Additionally, take a symmetric adjustment based on the mean of the retaining distribution (this step serves as an adaptive penalty), with a probability based on how likely $v$ is sampled from the forgetting distribution.
\end{enumerate}

\section{Feed-Forward Layers Key Space Hypercube Detection}
\label{sec:description_of_KSP_idea}
One limitation of \texttt{Neuron Adjust} is that it treats each neuron individually, without considering their correlations. However, neurons often work together in a coordinated way, contributing to the model's overall behavior. In this section, we first present our observation that query vectors evoking a specific skill tend to cluster in the key space of the feed-forward layers, as described in Section \ref{section:observation_clustering}. Building on this clustering phenomenon, we introduce \texttt{Key Space Detection (KSD)} in Section \ref{section:vector_pruning}, a machine unlearning technique that prevents query embedding vectors from accessing designated hypercubes.

\begin{figure}[t] 
  \includegraphics[width=\columnwidth] {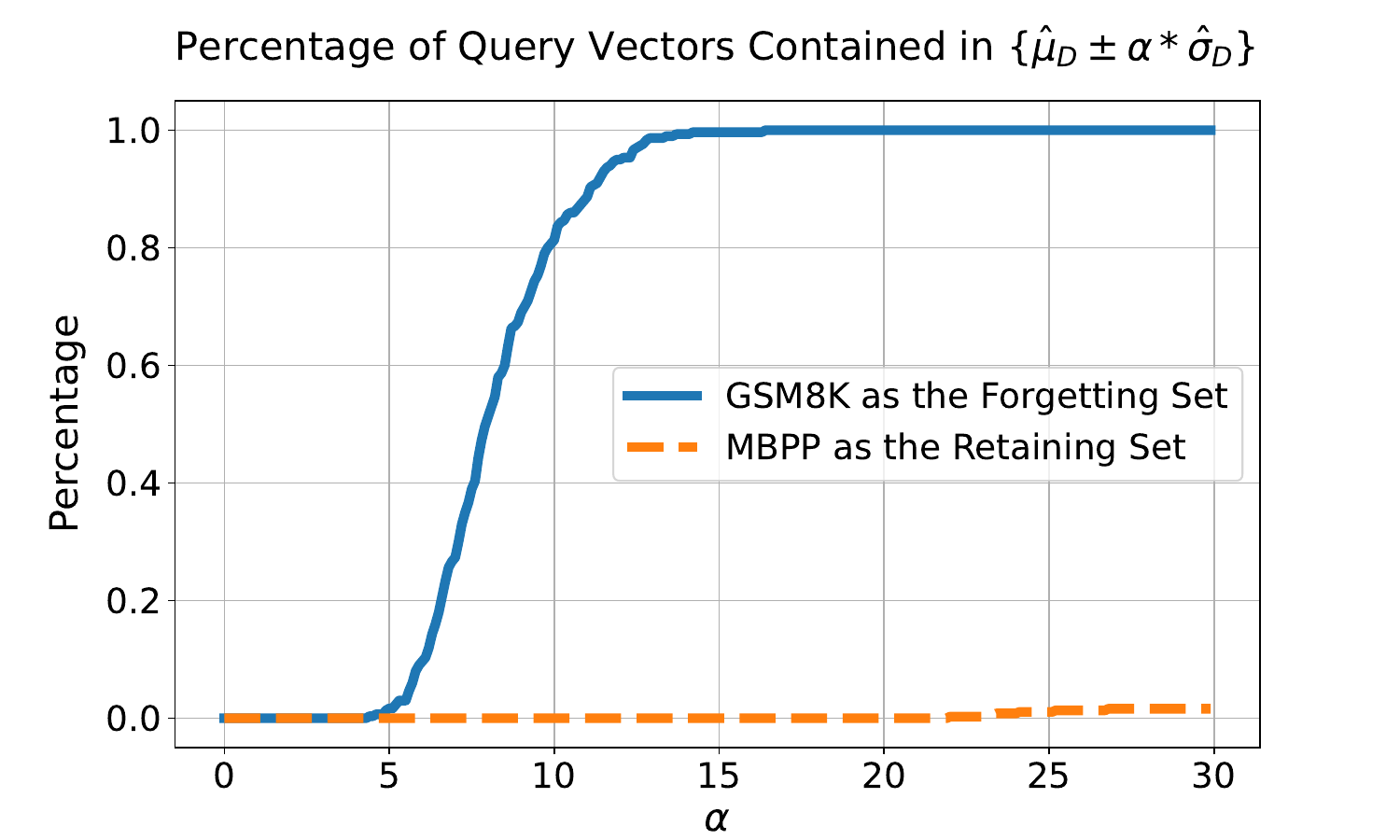}
  \vspace{-10pt}
  \caption{Percentage of Query Vectors contained in the Hypercube $\{\vec\mu_{\text{gsm8k}} \pm \alpha\vec\sigma_{\text{gsm8k}}\}$}
  \label{fig:hyperrectangle}
\end{figure}

\subsection{Neurons are Correlated Features in High-Dimensional Space} 
\label{section:observation_clustering}

\begin{figure*}[h!]
  \includegraphics[width=0.50\linewidth]{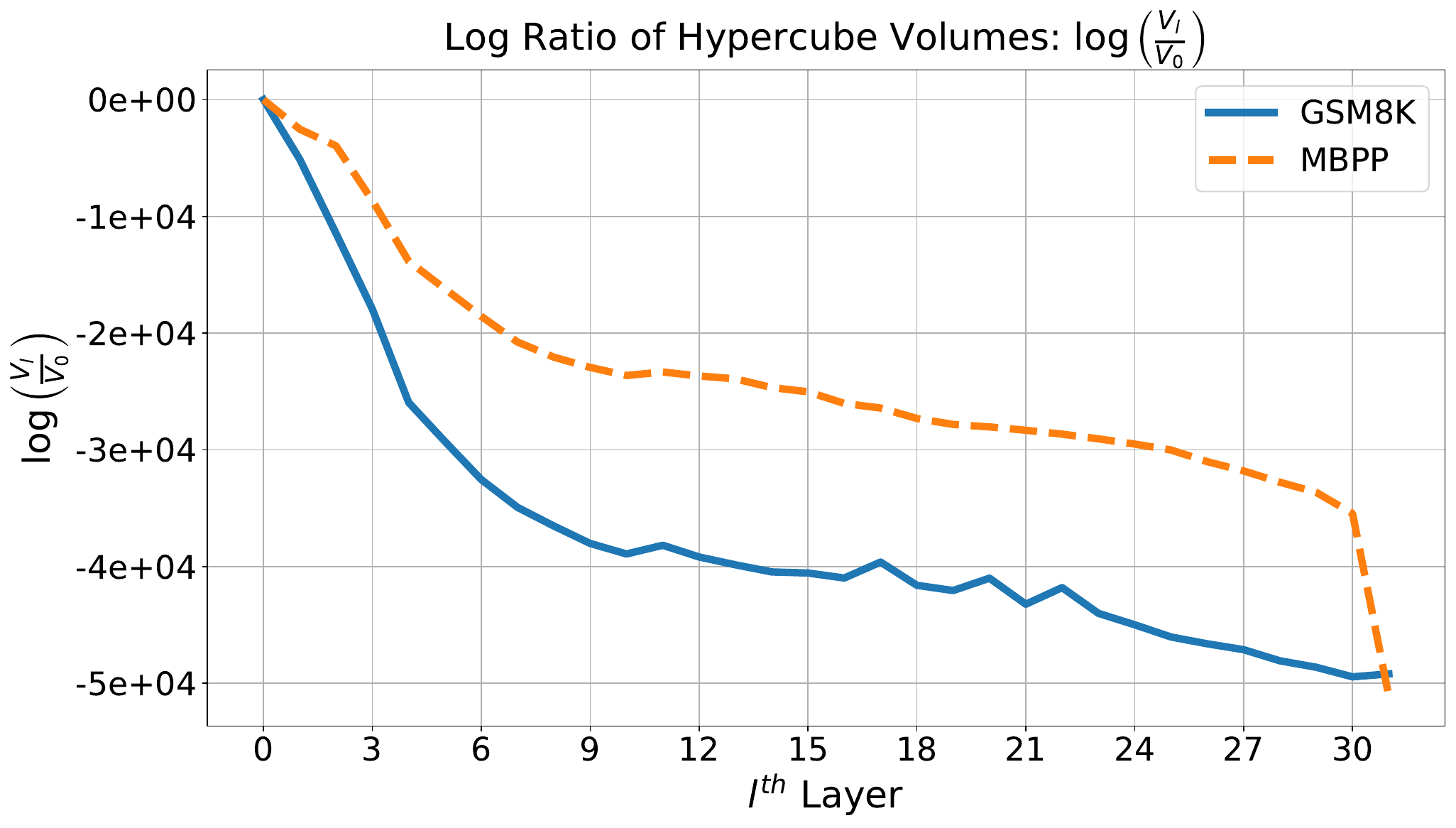} \hfill
  \includegraphics[width=0.475\linewidth]{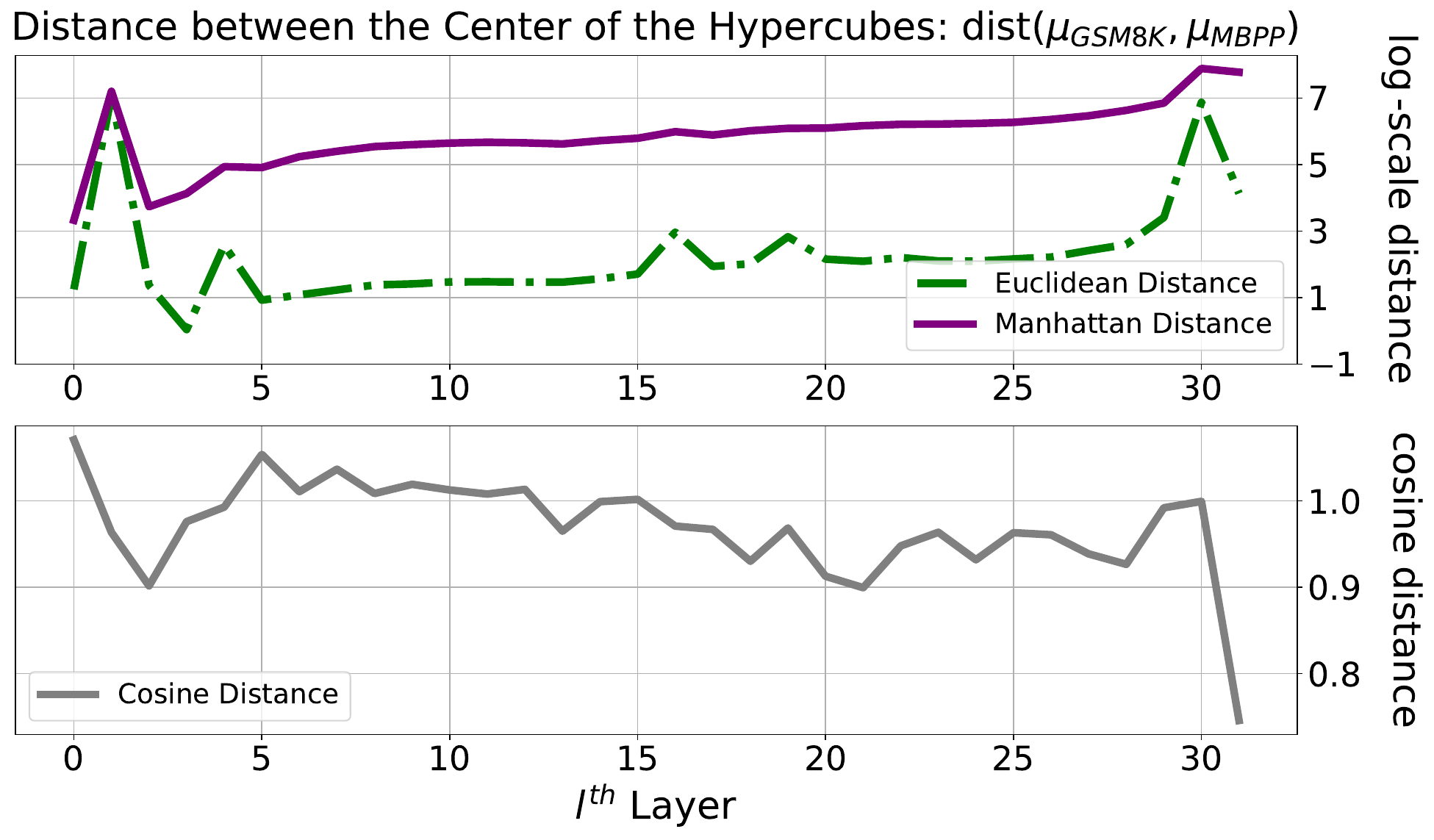}
  \vspace{-10pt}
  \caption {Relationship of the smallest hypercube containing all query vectors across layers. The left figure shows that different skill vectors cluster more tightly as the layer gets deeper. The right figure shows the distance between the centers of two hypercubes under different metrics.
  }
    \label{fig:clustering_distance_size}
\end{figure*}

As the $\vec{v}_{\text{key}}$ in Figure \ref{fig:MU_hyperrectangle}(b), we define the vector before the down-projection matrix $W_{\text{down}}$ in each FFL as the neuron activation vector, which forms the key space of each FFL. Each FFL has a unique key space. We use \texttt{llama-3-8b} as the subject language model, MBPP as the probing dataset that triggers the model's Python coding skill, and GSM8K as the probing dataset that triggers the model's grade school math problem-solving skill. We probe the model with the two training datasets and document two probing activation vector sets of the last token of each query, $\{\vec{v}^{(l)}\}_{\text{mbpp}}$ and $\{\vec{v}^{(l)}\}_{\text{gsm8k}}$, for the $l^{\text{th}}$ FFL. We calculate the mean and std vector, $\vec\mu_{D}^{(l)}$ and $\vec\sigma_{D}^{(l)}$, of each $\{\vec{v}^{(l)}\}_{D}$, where
$$
(\vec{\mu}_D^{(l)})_i \coloneqq \frac{1}{|D|}\sum_{j = 1}^{|D|}(\vec{v}_{j}^{(l)})_i
$$
$$
(\vec{\sigma}_D^{(l)})_i \coloneqq \sqrt{\frac{1}{|D|}\sum_{j = 1}^{|D|} \left( (\vec{v}_{j}^{(l)})_i - (\vec{\mu}_D^{(l)})_i \right)^2 },
$$
and bound the two vector sets with hypercubes 
$$
\{\vec\mu_{D} \pm \alpha\vec\sigma_{D}\}\!\coloneqq\!\{\vec{u}\;|\;\vec{\mu}_{D}^{(l)} - \alpha \vec{\sigma}_{D}^{(l)} \!\prec\!\vec{u}\!\prec\! \vec{\mu}_{D}^{(l)} + \alpha \vec{\sigma}_{D}^{(l)}\},
$$
where $\alpha$ is a hyperparameter that controls the size of the hypercube, and $\prec$ denotes element-wise less-than comparison. Figure \ref{fig:hyperrectangle} shows the percentage of vectors contained in the hypercube $\{\vec\mu_{\text{gsm8k}} \pm \alpha\vec\sigma_{\text{gsm8k}}\}$ when increasing $\alpha$ from $0$ to $30$ in the last layer of the model. We observe that when 
$\alpha = 15$, nearly all math query vector embeddings are encompassed within the hypercube. As $\alpha$ increases from $15$ to $20$, a gap forms between the math and code query clusters: all math queries remain within the hypercube, but no code queries are included. When $\alpha$ exceeds $20$, a few code queries begin to fall within the hypercube.

We further analyze the changes in size and distance within and between the math query cluster and the code query cluster across different layers. Specifically, for each layer $l$, we calculate the smallest hypercube that encompass all math and code queries, respectively, and compare their volumes and the distance between their centers. 

Figure \ref{fig:clustering_distance_size} (left) shows the log ratio of the volume of the $l^{\text{th}}$ layer hypercube to the volume of the first layer hypercube. As the layers get deeper, the hypercube becomes smaller, indicating denser clustering. Figure \ref{fig:clustering_distance_size} (right) shows the Euclidean, Manhattan, and cosine distances between $\vec{\mu}_{\text{gsm8k}}^{(l)}$ and $\vec{\mu}_{\text{mbpp}}^{(l)}$ for each $l$. We observe that the Euclidean and Manhattan distances gradually increase as the layers get deeper, except for high peaks in the very first and last layers. Additionally, for most layers, the cosine distance fluctuates around 1.0, indicating the orthogonality of the two clusters.

\begin{figure*}[t]
  \includegraphics[width=1\linewidth]{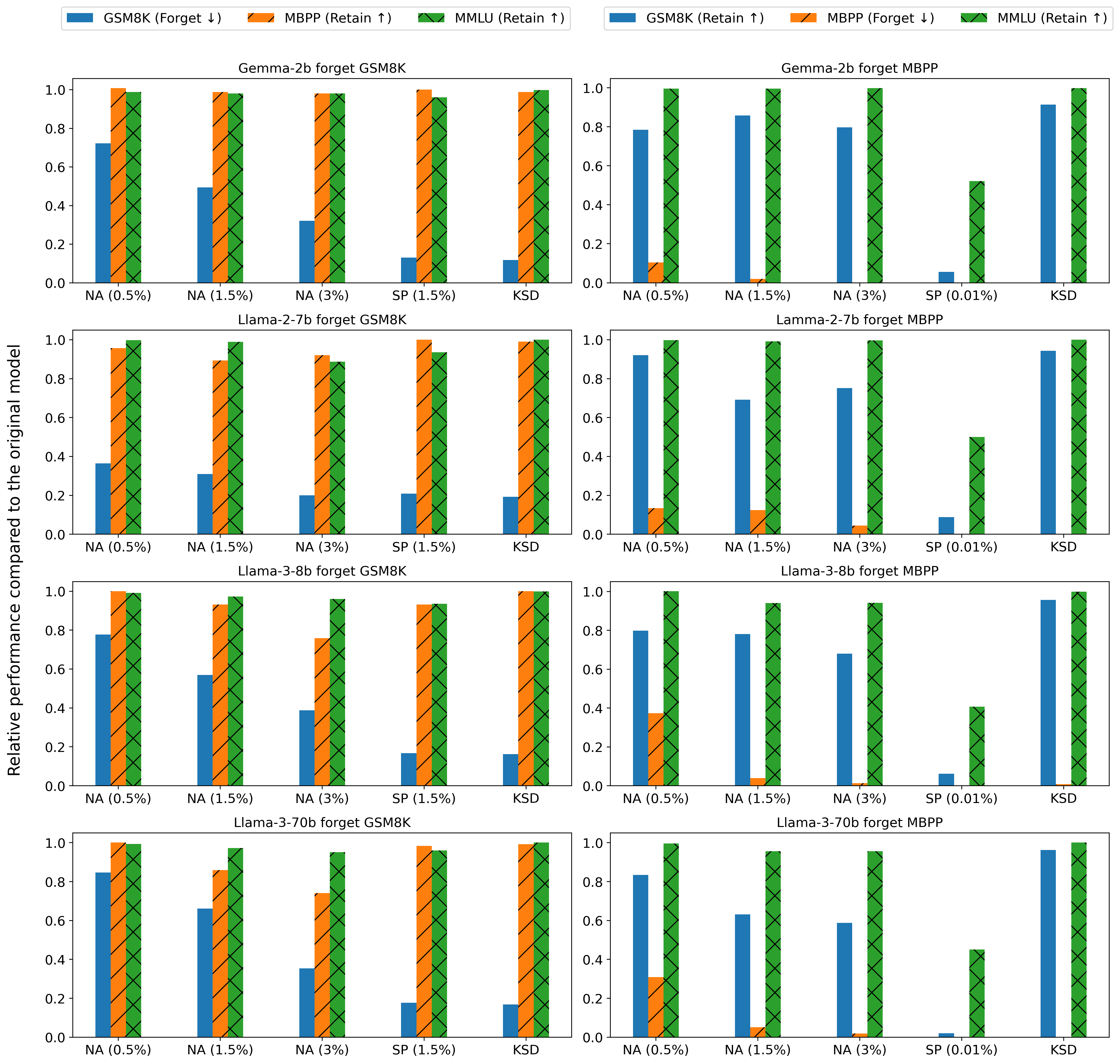}
  \vspace{-20pt}
  \caption {Performance of \texttt{Neuron Adjust} and \texttt{Key Space Detection} on Math/Code Skill Unlearning. On the horizontal axis, \texttt{NA}, \texttt{SP}, and \texttt{KSD} stand for \texttt{Neuron Adjust (ratio)}, \texttt{Selective Pruning (ratio)}, and \texttt{Key Space Detection}, respectively. The vertical axis represents the relative performance of the model compared to the original model after applying each unlearning method.}
  \label{fig:MathCode_Bar_plot}
\end{figure*}

\begin{figure*}[t!]
  \includegraphics[width=\columnwidth]
  {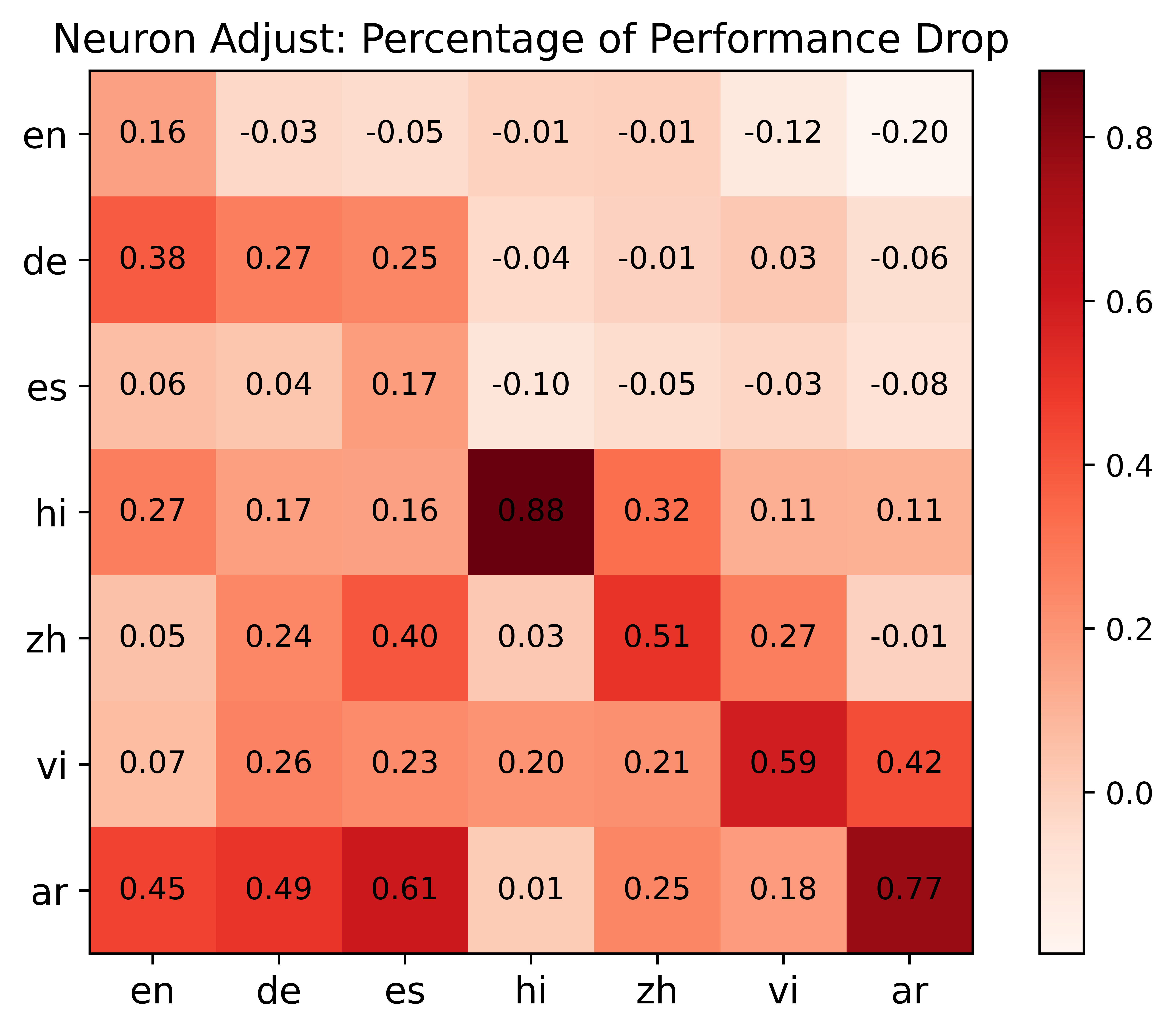} \hfill
  \includegraphics[width=\columnwidth]{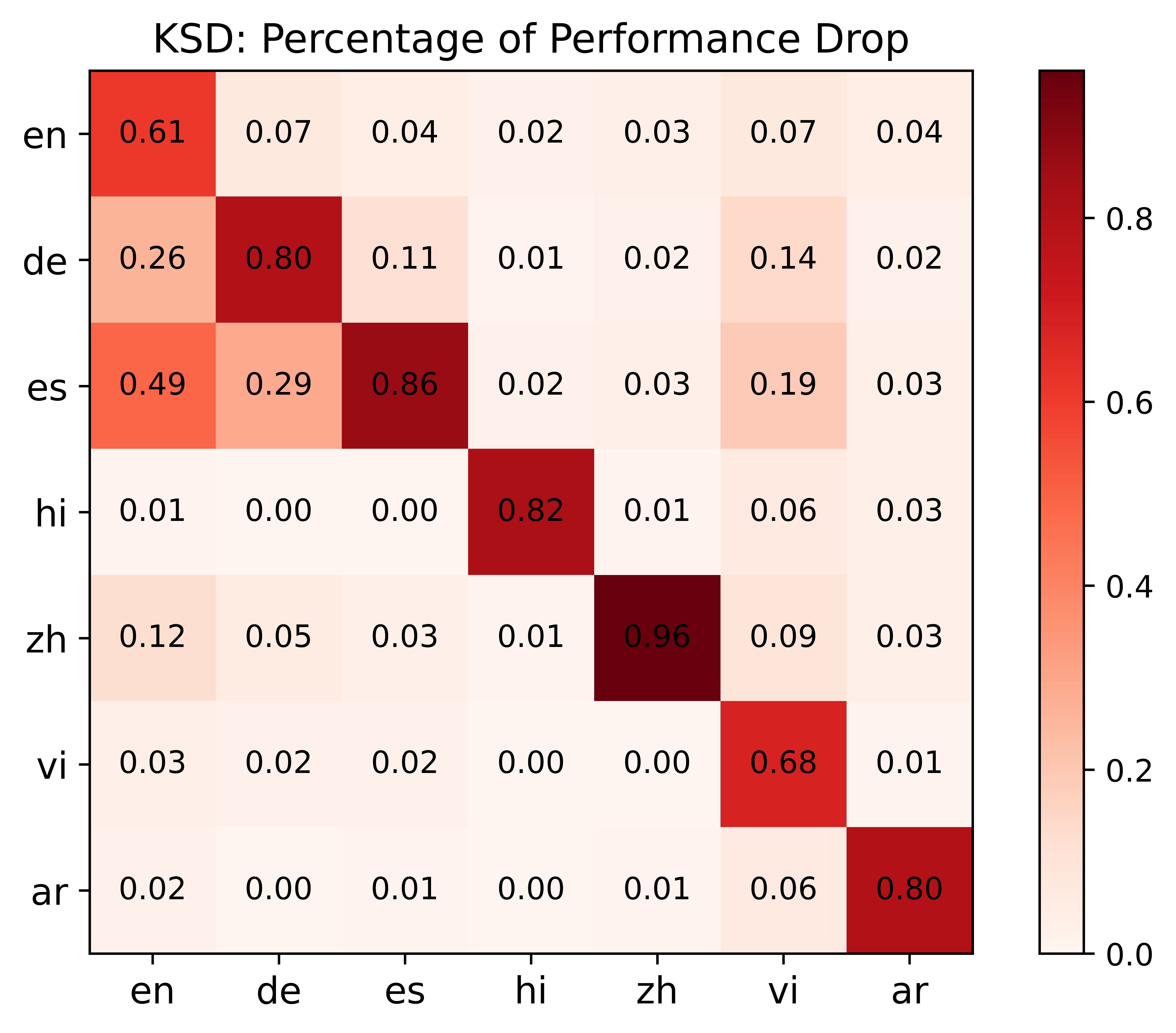}
  \vspace{-10pt}
  \caption {Results of unlearning one language in MLQA dataset while retaining the others with \texttt{Neuron Adjust 5\%} and \texttt{Key Space Detection} on \texttt{llama-3-8b}. \Revision{The i-th row shows the model's performance drop on different languages after unlearning the i-th language.}
  }
    \label{fig:MLQA_heat_map}
\end{figure*}

\begin{figure*}[h]
  \includegraphics[width=0.50\linewidth]{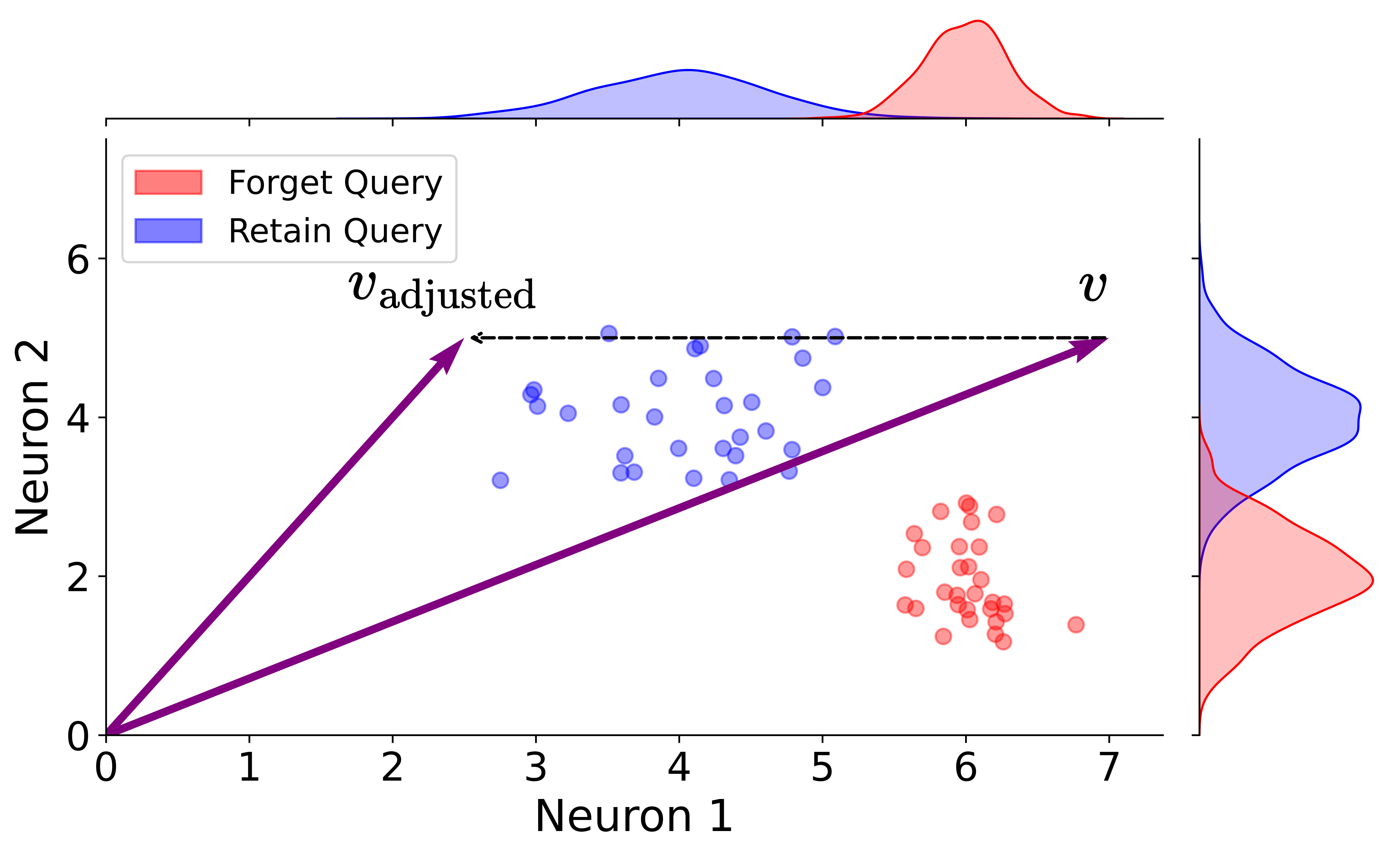} \hfill
  \includegraphics[width=0.50\linewidth]{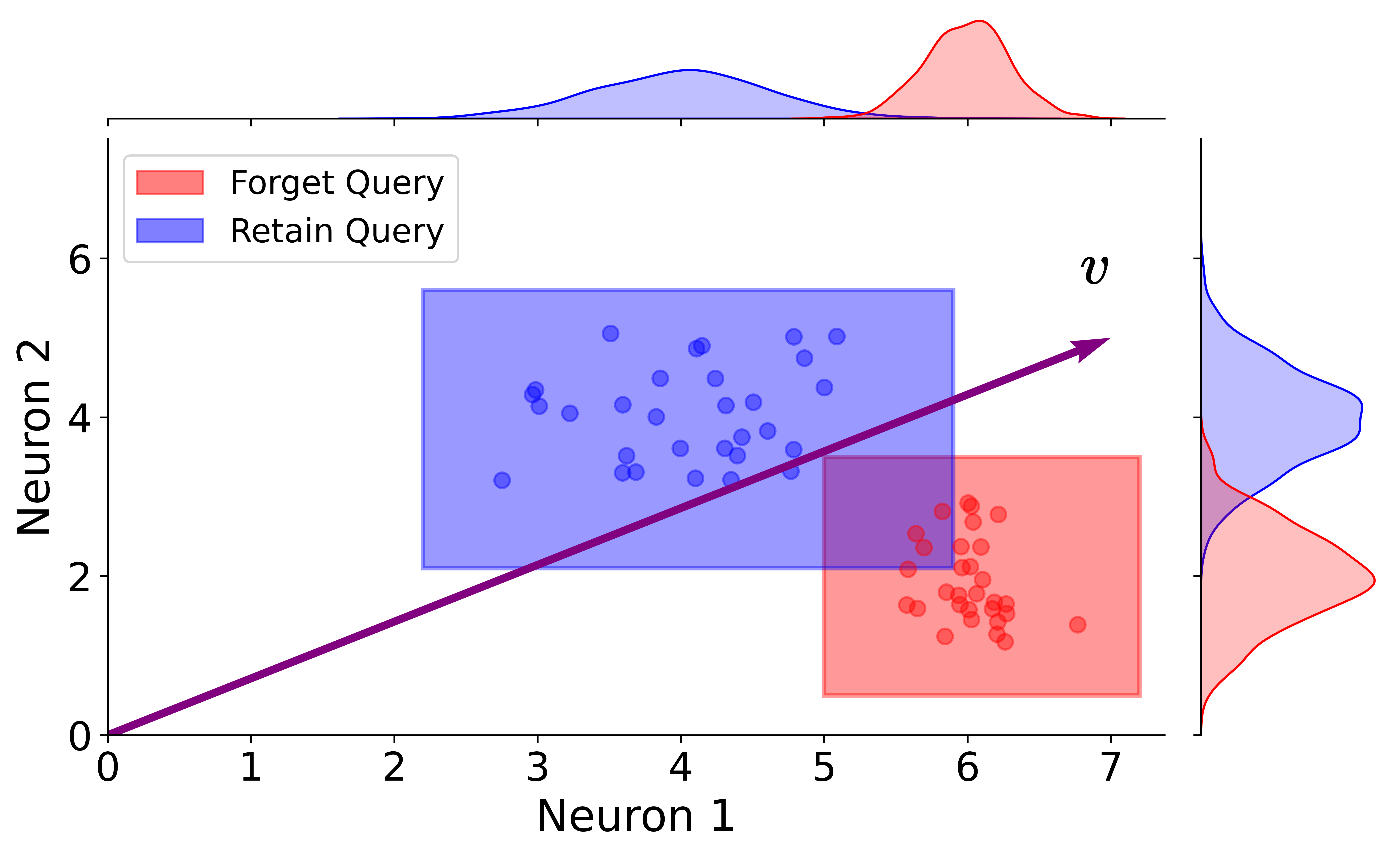}
  \vspace{-20pt}
  \caption {A case in a 2-neuron key space to explain how inference-time unrelated key vector would be affected by \texttt{Neuron Adjust} (left, get adjusted) and \texttt{Key Space Detection} (right, unaffected). \Revision{In this case, an adjustment to $v$ is unfavorable.}
  }
    \label{fig:KSD_NA_illustration}
\end{figure*}

\subsection{Machine Unlearning via Key Space Detection} \label{section:vector_pruning}

The idea of \texttt{Key Space Detection} is as follows: we first identify the sample mean vector $\vec{\mu}_{D}^{(l)}$ and the sample standard deviation vector $\vec{\sigma}_{D}^{(l)}$ of each FFL activation by probing the model with the forgetting dataset $D$. Then, we create a hypercube $$\{\vec\mu_{D}^{(l)} \pm \alpha\vec\sigma_{D}^{(l)}\}$$ in the key space as in section \ref{sec:description_of_KSP_idea}, where $\alpha$ is a hyperparameter we select to balance the trade-off between the quality of forgetting and maintaining the overall capability of the model. During inference, if we detect a query vector falling within the hypercube, we abstain the model's output and replace it with a system message "\texttt{Your query is not valid.}" instead.

\begin{algorithm}[h!]
\caption{\texttt{KSD} Algorithm for layer $l$}
\begin{algorithmic}[1]
\State \textbf{Input:} size hyperparameter $\alpha$, inference time activation key vector $v_{\text{key}}$, forgetting dataset $D$

\State Probing the model with $D$, estimate sample mean and std vectors $\vec{\mu}_{D}^{(l)}, \vec{\sigma}_{D}^{(l)}.$
\State During inference step $k$, current output $o$:
\If{$v_{\text{key}} \in \{\vec\mu_{D}^{(l)} \pm \alpha\vec\sigma_{D}^{(l)}\}$}
    \State $o \coloneqq$ "\texttt{Your query is not valid.}"
    \State Stop inference.
\Else 
    \State $o \text{ += } \text{token}_{k}$.
    \State Continue with the next token inference step \indent $k+1$.
\EndIf
\State \textbf{Output:} $o$
\end{algorithmic}\label{algo:ksd}
\end{algorithm}

\section{Experiments}

In this section, we present experimental results of math/code skill unlearning in Section \ref{sec:math_code_results} and language unlearning in Section \ref{sec:language unlearning}.  

We test the performance of \texttt{Neuron Adjust (NA)} and \texttt{Key Space Pruning (KSD)} on each skill unlearning task. For the \texttt{NA} method, we rank neurons by their difference in distribution mean, $\mu_f - \mu_r$, and select the top $\beta$ neurons with $\beta$ set to 0.5\%, 1.5\%, and 3.0\%. For the \texttt{KSD} method, we choose size coefficient $\alpha$ such that \texttt{KSD} either matches or outperforms the best forgetting quality of \texttt{Neuron Adjust}. \Revision{For the math/code skill unlearning task, we use \texttt{Selective Pruning} \cite{pochinkov2023dissecting}, a pruning-based skill unlearning method, as our baseline. This method prunes neurons in the FFLs based on their relative importance to each dataset.} After skill unlearning, we want the model to unlearn only the specific skill while maintaining its overall capabilities. Therefore, we also evaluate the effect of each method on 5-shot MMLU accuracy. For each experiment, we run it three times and report the best result. 

\subsection{Math/Code Skill Unlearning}
\label{sec:math_code_results}
For math/code skill unlearning, we choose the training splits of MBPP and GSM8K as the forgetting datasets. The MBPP dataset contains code-based programming problems for evaluating LLMs' Python code generation abilities, while the GSM8K dataset consists of grade school-level math problems for assessing their problem-solving skills in elementary mathematics. We use these two datasets to capture the model's Python coding skills and elementary math problem-solving skills. After unlearning, we test the models on the testing split of each dataset. For Python coding skills, we also test on MBPP+ \cite{evalplus}, which contains 35x more Python test cases to evaluate the robustness of the unlearning process.

Figure \ref{fig:MathCode_Bar_plot} compares the performance of the unlearning methods on four models: \texttt{Gemma-2b} \cite{gemmateam2024gemma}, \texttt{Llama-2-7b} \cite{touvron2023llama}, \texttt{Llama-3-8b}, and \texttt{Llama-3-70b} \cite{llama3modelcard}. \texttt{NA}, tested at different adjustment ratios, shows that higher ratios lead to more forgetting with minimal impact on MMLU accuracy, though MBPP retention decreases slightly. \texttt{KSD} achieves the highest forgetting rates with negligible effects on MMLU accuracy and retention, proving its efficiency in unlearning while maintaining overall performance. In contrast, \texttt{SP} effectively forgets math skills and retains coding performance but produces nonsensical output when asked to forget coding skills while retaining math, even with just $0.01\%$ neuron pruning. This suggests pruning can harm overall model capabilities, likely due to shared neurons between coding and math tasks. In this case, both \texttt{NA} and \texttt{KSD} outperform \texttt{SP}.

\subsection{Language Skill Unlearning}
\label{sec:language unlearning}
For the language skill unlearning task, we use the MLQA \cite{lewis2019mlqa} dataset as our evaluation benchmark. We use each language's context data as the forgetting dataset. Figure \ref{fig:MLQA_heat_map} shows the results of \texttt{NA} (left) and \texttt{KSD} (right) in a heatmap view. The $k^{\text{th}}$ row of the heatmap indicates the percentage decrease in each language's performance after forcing the model to forget the $k^{\text{th}}$ language on the vertical axis. Ideally, we aim to maximize the diagonal entries while keeping the other entries small. From the heatmap, we observe that \texttt{NA} performs well in forgetting English (en), Spanish (es), and Hindi (hi). However, when forgetting German (de), Chinese (zh), Vietnamese (vi), and Arabic (ar), the performance of one or more languages in the retaining dataset also decreases significantly. This suggests that the model may utilize a shared set of neuron values when demonstrating these languages. In contrast, \texttt{KSD} shows a better ability to maintain the model's performance on other languages while achieving substantial forgetting quality on most of the languages. After forgetting each language, we also tested the model's performance on the MMLU task. Both methods demonstrate a performance decrease of less than $5\%$. 
We also observe that \texttt{KSD} consistently outperforms other methods in retaining the model's overall capability. Figure \ref{fig:KSD_NA_illustration} illustrates the reason. For an out-of-distribution knowledge query vector $v$, \texttt{NA} may adjust some of its dimensions because it only considers operations for single neurons. In contrast, \texttt{KSD} considers the correlations among all neurons, prohibiting query vectors from accessing a much smaller area in the key space. Therefore, it is guaranteed to have no negative effect on out-of-hypercube queries.

\section{Conclusion}

In this paper, we propose two lightweight, training-free machine skill unlearning methods, \texttt{Neuron Adjust} and \texttt{Key Space Detection}, which have minimal impact on the model's overall capabilities. \texttt{Neuron Adjust} achieves unlearning by shifting the pre-activation of feed-forward layer neurons from the forgetting distribution to the retaining distribution. \texttt{KSD} achieves unlearning by prohibiting query vectors from accessing skill-specific key space. We evaluate our methods on unlearning math-solving, Python-coding, and comprehension skills across seven different languages with GSM8K, MBPP, and MLQA datasets respectively. Both methods show strong skill unlearning performance with minimal hurt to the model's overall capability. Experiments demonstrate the effectiveness of the two methods with > 80\% relative performance drop on the target forgetting skill and < 10\% drop on the model’s general knowledge and other skills. Specifically, Key Space Detection achieves nearly perfect skill unlearning with negligible overall capability drop.

Our findings provide insights into how neuron activations cluster in key spaces and how these spatial properties can be leveraged for skill unlearning. These observations not only enhance our understanding of model behavior but also offer a promising direction for more targeted and interpretable unlearning techniques. We believe these insights will contribute to ongoing investigations into model interpretability, safety, and control. For example, certain knowledge, such as personal privacy data or adversarial attack inputs, may be localized in more fine-grained regions of the model’s key space. Understanding how these spatial properties can be systematically utilized needs further exploration.

\section*{Acknowledgement}
The authors are partially supported by National Science Foundation under Grant No. 2107189, 2313105, 2430539, Hellman Fellowship, and Intel Rising Star Faculty Award. The authors would also like to thank anonymous reviewers for valuable feedback to improve the manuscript.

\section*{Limitations}

Similar to other unlearning methods, our approach can only remove capabilities that can be captured by a dataset. However, in real-world applications, a specific dataset may not always be available for every capability we wish to remove. Unlearning knowledge without a controlled dataset or unlearning out-of-distribution data points presents an interesting yet challenging problem in the field of machine unlearning. Future work could explore techniques to address this challenge.
Additionally, unlearning one skill while retaining a highly dependent skill requires a more fine-grained analysis. A promising direction for future research is to leverage spatial correlations among neurons to refine unlearning mechanisms and improve selectivity.
Furthermore, we have not yet identified an efficient and automatic way to determine the optimal values for the adjusting ratio and the size hyperparameter $\alpha$ in both methods. In the case of \texttt{Neuron Adjust}, the reduction of unintended capabilities in the model is not guaranteed. For \texttt{Key Space Detection}, although we can ensure the model's performance for out-of-hypercube queries, it may still lead to the degradation of certain unknown capabilities, as queries may cluster in a non-convex shape within the key space.

\bibliography{custom}

\appendix
\onecolumn
\addcontentsline{toc}{section}{Appendix} 
\part{} 
\parttoc 
\section{Appendix}
\label{sec:appendix}
\subsection{Time complexity analysis}
\texttt{Neuron Adjust} and \texttt{KSD} are both plug-in modules applicable to any autoregressive LLMs. For both methods, the implementation involves obtaining the mean and standard deviation of each neuron in the key space for every MLP layer by probing the subject model with the forgetting dataset. Since the forgetting dataset consists of $N$ samples, this requires only $N$ forward passes of the subject model. Given an autoregressive LLM with $L$ MLP layers, each containing $K$ neurons (where $L$ and $K$ are constants), we analyze the time complexity in detail:

\paragraph{\texttt{Neuron Adjust:}} For each inference step, we need to:

\begin{itemize}
    \item Determine whether the inference time neuron activation is more likely to be drawn from the forgetting or retaining distribution. This step is $O(KL)=O(1)$.
    \item Change the neuron activation value if necessary. This step is also $O(KL)=O(1)$.
\end{itemize}
Therefore, Neuron Adjust has an inference time cost of $O(1)$.

\paragraph{\texttt{KSD}:} For each inference step, we only need to determine whether the key vector in the last MLP layer is within the forgetting hyper-rectangle. This step is $O(L)=O(1)$. Therefore, \texttt{KSD} has an inference time cost of $O(1)$.
\newline
\newline
\noindent
In our experiments (real-world setting), it took less than 15 mins to implement our methods on llama-3-8b with a single V100 GPU, thus they are pretty light compared to other training-based methods.

\newpage

\subsection{Sequential forgetting of multiple skills}
In this section we present the behavior of our methods when tasked to forget multiple skills sequentially. As shown in section \ref{sec:neuron_adjust} and \ref{sec:description_of_KSP_idea}, our methods are designed to be applied as plug-in modules. These modules can be used after each model update or fine-tuning process without affecting the model’s ability to learn new tasks or skills.

\texttt{Neuron Adjust} is specifically optimized for forgetting a single skill, while \texttt{Key Space Detection (KSD)} can be extended to forget multiple skills. \texttt{KSD} works by identifying the hypercube that corresponds to each skill and determining whether the inference-time key vector falls within any of these hyper-rectangles. The computational complexity of this approach at inference time is $O(M)$, where $M$ represents the number of skills to be forgotten. 

To demonstrate the effectiveness of \texttt{KSD} in forgetting multiple skills, we conducted an additional experiment using \texttt{Llama-3-70B}, targeting the simultaneous forgetting of two tasks, MBPP and GSM8K. The results, as shown in Table \ref{tab:forgetting_results}, highlight the significant reduction in performance on the forgotten skills while maintaining general MMLU performance.

\begin{table}[H]
\centering
\resizebox{0.6\columnwidth}{!}{ 
\begin{tabular}{|l|c|c|c|c|}
\hline
Method & GSM8K & MBPP & MBPP+ & MMLU \\ \hline
Original & 47.5\% & 61.1\% & 51.1\% & 64.9\% \\ \hline
\texttt{KSD} & 8.1\% & 0.5\% & 0.5\% & 64.8\% \\ \hline
\end{tabular}
}
\caption{Results of Forgetting MBPP and GSM8K with Llama-3-70B}
\label{tab:forgetting_results}
\end{table}

The results demonstrate that \texttt{KSD} achieves a performance drop of over 80\% on both GSM8K and MBPP tasks, effectively erasing the learned skills while leaving general knowledge tasks largely unaffected.

\end{document}